\documentclass[letterpaper, 10 pt, conference]{ieeeconf}  
\IEEEoverridecommandlockouts                              

\usepackage{graphics} 
\usepackage{epsfig} 
\usepackage{mathptmx} 
\usepackage{times} 
\usepackage{amsmath} 
\usepackage{amssymb}  
\usepackage[labelformat=simple]{subcaption}
\usepackage{array}

\usepackage{dblfloatfix}

\usepackage{xspace}
\makeatletter
\DeclareRobustCommand\onedot{\futurelet\@let@token\@onedot}
\def\@onedot{\ifx\@let@token.\else.\null\fi\xspace}
\def\eg{\emph{e.g}\onedot}

\makeatother

\newcommand{\specialcell}[2][c]{%
  \begin{tabular}[#1]{@{}c@{}}#2\end{tabular}}

\title{\LARGE \bf
SHOP-VRB: A Visual Reasoning Benchmark for Object Perception
}

\author{Michal Nazarczuk$^{1}$ and Krystian Mikolajczyk$^{1}$
\thanks{$^{1}$Authors are with Department of Electrical and Electronic Engineering,
        Imperial College London, London SW7 2AZ, United Kingdom
        {\tt\small [michal.nazarczuk17, k.mikolajczyk]@imperial.ac.uk}}%
}

\usepackage{hyperref}

\begin{document}

\maketitle
\thispagestyle{empty}
\pagestyle{empty}

\begin{abstract}
   In this paper we present an approach and a benchmark for visual reasoning in robotics applications, in particular small object grasping and manipulation. The approach and benchmark are focused on inferring object properties from visual and text data. It concerns small household objects with their properties, functionality, natural language descriptions as well as question-answer pairs for visual reasoning queries along with their corresponding scene semantic representations. We also  present a method for generating synthetic data which allows to extend the benchmark to other objects or scenes and propose an evaluation protocol that is more challenging than in the existing datasets. 

    We propose a reasoning system based on symbolic program execution.
    A disentangled representation of the visual and textual inputs is obtained and used to execute symbolic programs that represent a 'reasoning process' of the algorithm. 
    We perform a set of experiments on the proposed benchmark and compare to results from the state of the art methods. These results expose the shortcomings of the existing benchmarks that may lead to misleading conclusions on the actual performance of the visual reasoning systems.
    
\end{abstract}

\section{Introduction}

Visual Reasoning or Visual Question Answering address the problem of correctly interpreting a visual input with a natural language question, and perform logical reasoning that results in the correct answer. One of the approaches to perform such reasoning is via creating a symbolic program and  executing it on the representation of the observed scene. This approach was initially investigated by CLEVR-IEP \cite{Johnson2017InferringReasoning} but abandoned due to limited performance. Its new implementations  have recently emerged showing impressive performance \cite{Mao2019TheSupervision, Yi2018Neural-SymbolicUnderstanding}. We note a striking resemblance of the symbolic program execution to the task of robotic perception, i.e. to analyse visual cues of the environment and perform a set of logical actions to achieve a goal. This can be interpreted as answering a specific question on an observed image that requires reasoning.

Benchmarking in robotics is a challenging task due to different
physical robotic platforms, test sets, evaluation scenarios, or simulators; the criteria for a
correct grasp is often only binary i.e. success/failure etc. 
The best practice so far is represented by YCB \cite{Calli2015TheResearch},
which provides physical objects, detailed descriptions of the experimental protocol, and clear benchmarking metrics.

\begin{figure}[h]
    \centering
    \begin{subfigure}[b]{0.235\textwidth}
        \includegraphics[width=0.99\textwidth]{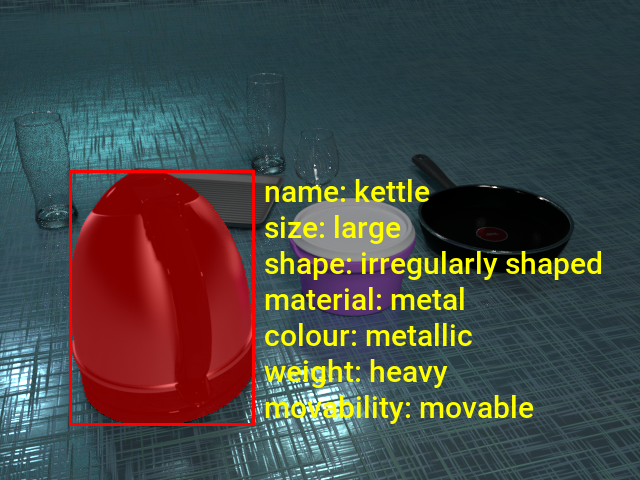}
        \caption{How many other things are the same color as the medium-sized light plastic object?}
        \label{fig:1a}
    \end{subfigure}
    \begin{subfigure}[b]{0.235\textwidth}
        \includegraphics[width=0.99\textwidth]{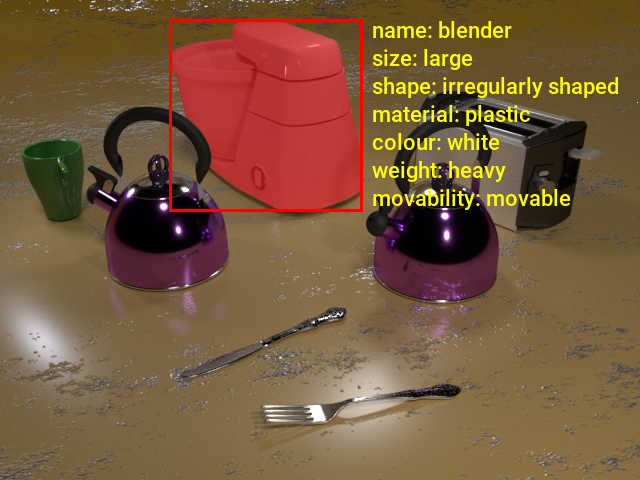}
        \caption{There is a large white plastic thing to the left of the toaster; what category does it belong to?}
        \label{fig:1b}
    \end{subfigure}
    \begin{subfigure}[b]{0.47\textwidth}
        \centering
        \includegraphics[width=0.49\textwidth]{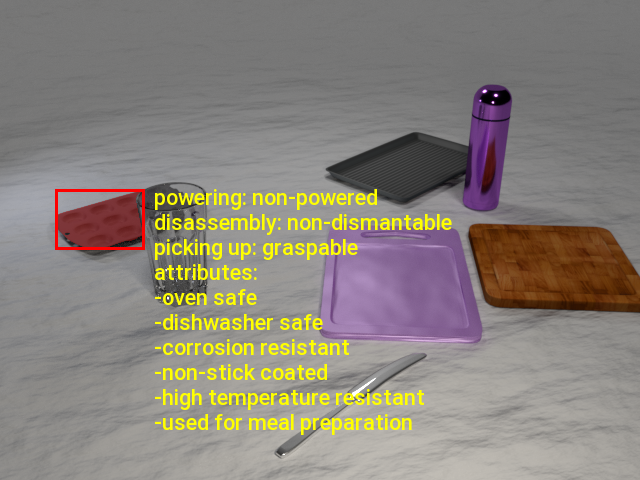}
        \parbox[b]{0.49\textwidth}{
            \footnotesize{No matter what you're cooking, this Oven Tray is suitable for the job. Whether you're cooking meats, veg or cookies - this oven tray will perform perfectly. The non-stick coating makes sure your treats come out in one piece, an easy grip makes for safe oven removal, plus it's dishwasher safe.}
        }
        \caption{How to move the object that is to the left of the purple plastic flat thing?}
        \label{fig:1c}
    \end{subfigure}
    \caption{Sample scenes from the dataset along with examples of corresponding questions of different types. Splits: \subref{fig:1a} train,  \subref{fig:1b} benchmark {(both presenting only visual properties)}, \subref{fig:1c} test ({only} textual properties). \textit{Test} split contains new scenes generated from known objects' instances (present in \textit{training} and \textit{validation}), whereas \textit{benchmark} is composed of new instances of the objects belonging to the known categories.} \label{fig:1}
\vspace{-2em}
\end{figure}

In this paper, we present an approach and a benchmark SHOP-VRB\footnote[2]{Available at: \url{https://michaal94.github.io/SHOP-VRB}} (Simple Household Object Properties) to bridge the gap between the requirements in robotic perception tasks and typical problems in visual reasoning. Figure \ref{fig:1} presents an example scene from the proposed benchmark. A benchmark needs to be challenging in terms of  object perception as well as scene composition, that should be interpretable and require complex reasoning for visual question answering. 

Therefore, we provide a large number of scenes, that are generated in a procedural way,  contain  household objects and appliances suitable for robotic grasping and manipulation. 
To improve upon the YCB benchmarking strategy, we 
include a wide range of object types and test scenarios with unambiguous descriptions of
experimental setups. 
We emphasise the importance of the choice of the experimental setup in order to better simulate real  conditions for assessing the generalisation potential of the proposed models, thus providing a suitable benchmark for the task. We focus our attention on methods that allow to obtain fully interpretable scene representation, on a human level of abstraction. We consider such methods suitable for real world applications that include human-robot knowledge exchange. 

In addition, each object category is provided with various textual descriptions which we believe will greatly benefit robotic perception. Once the object category is recognised, it is straightforward to find its various textual descriptions on the Internet (\eg \textit{Wikipedia}, manuals). Our approach makes use of such descriptions to infer object properties that cannot be extracted from visual data. This allows to infer non-visual attributes of objects without explicit training from labelled examples. We also make available the implementation of our method that can recognise and reason about
object properties.

\section{Related work}


Our work is inspired by YCB \cite{Calli2015TheResearch} and CLEVR \cite{Johnson2017CLEVR:Reasoning}, which are the  datasets for object grasping and for visual reasoning, respectively. YCB is well known in the robotics community and consists of real physical objects,  most of them rigid and Lambertian, their RGB-D
images and 3D meshes. CLEVR is mainly used by VQA systems and it is composed of synthetic scenes including simple geometrical objects such as spheres, cubes, or cylinders in various sizes, colours and materials. It includes question-answer pairs, as well as a graph of functional programs that can be used to generate answers. 
While YCB is widely used in robotics for object modelling and grasping, it does not pose a significant challenge in terms of computer vision analysis. Similarly, CLEVR was designed to benchmarking visual reasoning rather than visual perception which lead to a saturation of results, with the top accuracy of $99.8\%$ by NS-VQA \cite{Yi2018Neural-SymbolicUnderstanding}. We argue that it results from the fact that CLEVR and other similar datasets (CLEVR-Ref+ \cite{Liu2019CLEVR-Ref+:Expressions}, CLEVR-Human \cite{Johnson2017InferringReasoning}, Sort-of-CLEVR \cite{Santoro2017AReasoning}) are all composed of very simple objects and their composition in clear background resulting in nearly perfect perception accuracy.  We aim to address this issue with the proposed dataset.

In the context of robotics applications and object grasping/manipulation, the datasets usually focus  on the position and the grasping method of the object. In addition to YCB \cite{Calli2015TheResearch}, DexNet (v4.0) \cite{Mahler2019LearningPolicies} provides synthetically generated point clouds  annotated with suction and grasp force for robotic arm manipulation. 
{
ShapeNet \cite{Chang2015ShapeNet:Repository} provides a rich library of various 3D models. 
Relevant datasets are also offered in VQA field \cite{Goyal2017, Gurari2018VizWizPeople, Krishna2017VisualAnnotations, Hudson2019CompostionalReasoning}, which typically include  diverse scenes.
}


SHOP-VRB attempts to bridge the gap between VQA and robotics dataset. 
It is inspired by the idea of exploiting functional programs for visual reasoning. 
The concept of creating programs for VQA was presented in CLEVR-IEP \cite{Johnson2017InferringReasoning}, which uses module neural network as the executor evaluated on a latent representation obtained from CNN. Our proposed approach is  based on NS-VQA  \cite{Yi2018Neural-SymbolicUnderstanding}  that consists of  object segmentation, attributes extraction, question parsing and program execution. 
The method for extracting visual object properties is based on neural scene de-rendering \cite{Wu2017NeuralDe-rendering} which tries to obtain a structured and disentangled representation of the visual input.

There are other models for visual reasoning that explore recurrent approaches for scene processing, relation reasoning, module networks, or attention structures \cite{Hu2017LearningAnswering, Hudson2018CompositionalReasoning, Santoro2017AReasoning, Perez2018FiLM:Layer, Zhu2017StructuredAnswering}.
Some of these methods incorporate a logic structure of the reasoning task to the  model, and employ underlying functional programs for the questions \cite{Hu2018ExplainableNetworks, Mascharka2018TransparencyReasoning, Suarez2018DDRprog:Programmer, Shi2019ExplainableGraphs}. For example, NS-CL \cite{Mao2019TheSupervision} explores the possibility of executing programs on latent space image representation with the use of neuro-symbolic reasoning module.

\section{Dataset}


\begin{figure*}[h]
    \centering
    \includegraphics[width=\textwidth]{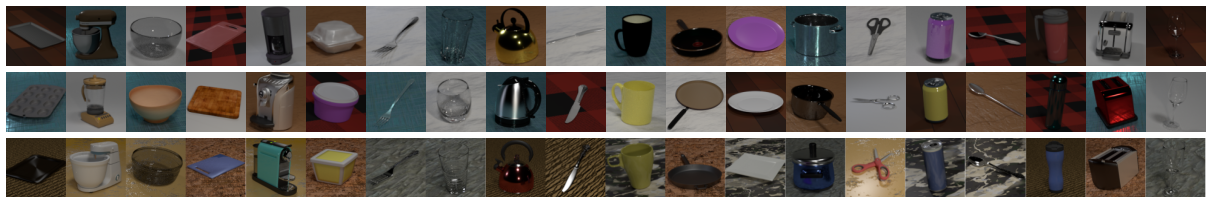}
    \caption{Examples of objects from SHOP-VRB - different instances, materials, colour and background settings. Items from first two rows can be found in \textit{training}, \textit{validation} and \textit{test} splits. Last row presents samples \textit{exclusive} for \textit{benchmark} subset.}\label{fig:exampl_shop}
    \vspace{-2mm}
\end{figure*}

SHOP-VRB (Simple Household Object Properties) provides a benchmark for visual reasoning and recovering structured, semantic representation of a scene. 
 In contrast to YCB  \cite{Calli2015TheResearch} and CLEVR \cite{Johnson2017CLEVR:Reasoning}, SHOP-VRB provides scenes with various kitchen objects and appliances, including articulated ones, along with questions associated with those scenes. Images of three instances of each of the 20 object classes can be seen in Figure \ref{fig:exampl_shop}. 
 Objects are represented by $66$ 3D models, $20$ of which are used  exclusively for \textit{benchmark} split. Each object class has $1$ to $5$ different instances in order to diversify their possible attributes and thus avoid overfitting to a specific shape. All object classes are listed in Table \ref{tab:objects} along with their number of instances. 
The list of attributes for each object includes:
\begin{itemize}\vspace{-\topsep}
\setlength\itemsep{-0.01em}
\small
    \item name - one of $20$ categories summarised in the Table \ref{tab:objects},
    \item size - \textit{small}, \textit{medium}, \textit{large}, 
    \item weight - \textit{light}, \textit{medium}, \textit{heavy},
    \item material - \textit{rubber}, \textit{metal}, \textit{plastic}, \textit{wood}, \textit{ceramic}, \textit{glass},
    \item colour -  \textit{gray}, \textit{red}, \textit{blue}, \textit{green}, \textit{brown}, \textit{purple}, \textit{cyan}, \textit{yellow}, \textit{white}, \textit{metallic}, \textit{transparent}, \textit{black},
    \item shape - \textit{cuboid}, \textit{irregularly shaped}, \textit{hemisphere}, \textit{cylindrical}, \textit{long and thin shaped}, \textit{flat},
    \item mobility - \textit{portable} (easily picked up), \textit{movable} (moved without picking up).
\end{itemize}\vspace{-\topsep}


\begin{table}[h]
\small
\begin{center}
\begin{tabular}{@{}llllll@{}}
\hline
Object name    & $\#$ & Object name & $\#$ & Object name & $\#$\\
\hline
Baking tray    & 2   & Glass       & 5   & Scissors    & 2 \\
Blender        & 2   & Kettle      & 2   & Soda can    & 1 \\
Bowl           & 3   & Knife       & 2   & Spoon       & 2 \\
Chopping board & 3   & Mug         & 3   & Thermos     & 2 \\
Coffee maker   & 2   & Pan         & 2   & Toaster     & 2 \\
Food box       & 2   & Plate       & 2   & Wine glass  & 3 \\
Fork           & 2   & Pot         & 2   & & \\
\hline
\end{tabular}
\end{center}
\caption{Objects represented in SHOP-VRB along with the number of instances (training, validation and test split).} \label{tab:objects}
\end{table}

Items in SHOP-VRB are assigned with material commonly used for such objects. Some instances may also be rendered with several different materials, \eg rubber or plastic chopping board, thus require the method to perform  material recognition during scene parsing. Subsequently, some materials (rubber, metal, plastic, ceramic) can be assigned  a random colour. Furthermore, different instances of objects appear in different sizes \eg smaller and bigger glass. Models for different object instances are chosen to include different shapes \eg cylindrical or irregular lunch box. 
Additionally, all classes are provided with $11$ short natural language descriptions allowing for questions to be grounded not only in the visual input but also requiring an external (textual) source of knowledge. 
Text descriptions extend the list of properties of each object by the following:
\begin{itemize}\vspace{-\topsep}
\setlength\itemsep{-0.01em}
\small
    \item powering - \textit{socket-powered}, \textit{non-powered},
    \item disassembly - \textit{non-dismantable}, \textit{unscrewable}, \textit{dismantable}, 
    \item picking up - \textit{handle-pickable}, \textit{graspable}, \textit{stationary},
    \item attributes (multi-label) - list of attributes from $36$ different values (\eg \textit{operated with buttons}, \textit{corrosion resistant}, \textit{having removable parts}), 
\end{itemize}

Scenes of SHOP-VRB are rendered with a set of different backgrounds characterised by different colours, textures and parameters of light diffusion.
Example scenes from the dataset are presented in Figure \ref{fig:1}. %
Each image in the dataset is associated with ground truth segmentation mask, object locations and a list of their attributes.

For the reasoning task, each scene is provided with $19$ or $20$ questions related to the presence of certain objects in the image, their count, integer comparison between objects, attributes comparison or querying for an attribute of an object. Additional $9$ to $10$ questions relate to the information required to be obtained from both visual and natural language description. Questions are provided with answers and functional programs that can be executed on structured scene representation \cite{Yi2018Neural-SymbolicUnderstanding}.

\subsection{Generation procedure}

SHOP-VRB is a synthetic dataset generated procedurally to allow for easy modifications, extensions and new data generation. 
\paragraph{Scenes} are generated on one of $6$ possible backgrounds. 
 The number of objects in the scene varies from $3$ to $7$, which are placed {on the horizontal supporting surface} one by one with a randomly selected 3D model. 
Intersections and significant occlusions are analysed and removed {by regenerating the scene if the number of visible pixels per object is lower than a threshold}.
%
All objects are assigned random material and colour out of the possible 
values.
 The scene is generated with ground truth annotations and segmentation mask.
\paragraph{Questions} are generated based on 488  
different templates i.e. 293 for visual only data, and 195 for both visual and textual properties. 
Each template consists of several phrasings of a question, a program leading to the answer and set of included or excluded parameter types, \eg when asking about the colour of the object, it is excluded from appearing in the question. For each scene, the program attempts to generate a  number of questions ($20$ for visual and $10$ for textual properties). Firstly, a random question template is chosen making sure the distribution of templates is uniform. Afterwards, a depth-first search is used to instantiate the question with parameters that require the question to have an answer and avoid ambiguity.

\subsection{SHOP-VRB splits}


 We provide two test splits, namely, \textit{test} and \textit{benchmark} to prevent quick performance saturation, such as on CLEVR data. The \textit{training}, \textit{validation}, and \textit{test}  subsets contain the same shapes of object instances but with different attributes, whereas \textit{benchmark} is composed of different shape instances to the other splits. SHOP-VRB \textit{benchmark} focuses on the practical scenario which requires autonomous systems to generalise to new instances of objects of similar functionality. 



Table \ref{tab:stats} shows details for all SHOP-VRB splits including the number of scenes and questions along with their uniqueness and overlap between subsets. Note that the scenes and questions are generated randomly according to simple rules therefore repetition is possible. 
Overlaps between sets do not exceed $1.5\%$. That provides a great diversity for the dataset and does not allow for biases that could be exploited by learning-based reasoning.

\newcommand{\centered}[1]{\begin{tabular}{c} #1 \end{tabular}}

\begin{table}[ht]
\small
\begin{center}
\begin{tabular}{@{}>{\centering\arraybackslash}m{0.10\textwidth}>{\centering\arraybackslash}m{0.06\textwidth}>{\centering\arraybackslash}m{0.08\textwidth}>{\centering\arraybackslash}m{0.06\textwidth}>{\centering\arraybackslash}m{0.09\textwidth}@{}}
\hline
    & Train & Validation & Test & Benchmark \\
\hline
Images              & 10000     & 1500      & 1500      & 1500    \\
Questions           & 199952 \textit{100000}    & 30000 \textit{15000}     & 30000 \textit{15000}    & 30000 \textit{15000}  \\
Unique questions    & 196769 \textit{95894}   & 29917 \textit{14727}    & 29908 \textit{14762}    & 29915 \textit{14785}  \\
Overlap with train  & -     &  \specialcell{873 \\ \textit{581}}     &  \specialcell{855 \\ \textit{608} }    &  \specialcell{676 \\ \textit{603} }  \\
Overlap with benchmark  &  \specialcell{676 \\ \textit{603} } &  \specialcell{117 \\ \textit{199} }    &  \specialcell{135  \\ \textit{223} }    & -    \\
\hline
\end{tabular}
\end{center}
\caption{Number of data samples in SHOP-VRB dataset (visual and \textit{textual} questions).} \label{tab:stats}
\vspace{-2mm}
\end{table}

\section{Model}
\begin{figure*}[h]
     \centering
     \includegraphics[width=0.9\textwidth]{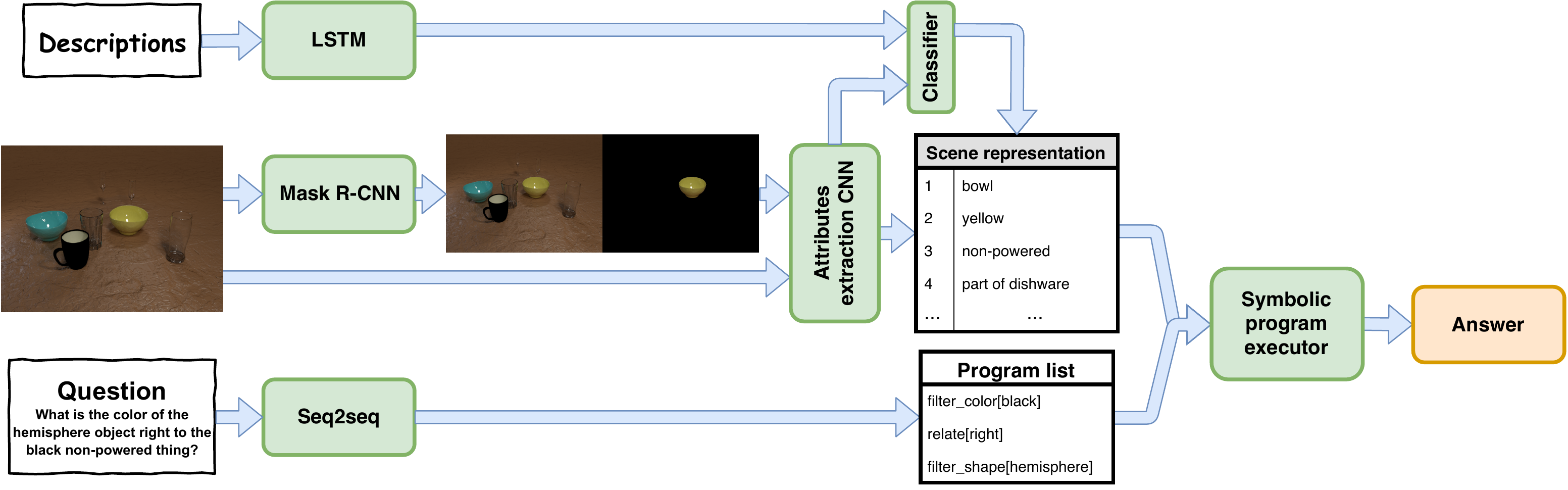}
     \caption{Our approach for visual reasoning. An image is parsed through Mask R-CNN to obtain object proposals which are used to extract disentangled scene representation. With seq2seq model a question is transformed into symbolic programs that are executed on the attribute representation in order to obtain the final answer.}\label{fig:model}
     \vspace{-3mm}
 \end{figure*}

%
We extend NS-VQA \cite{Yi2018Neural-SymbolicUnderstanding} approach to address the visual reasoning questions using image and text description. 
%
The method consists of four main components, namely, visual scene parser, textual description parser, command parser, and programme executor. The approach is illustrated in Figure \ref{fig:model}.

\paragraph{Scene parsing} Mask R-CNN \cite{He2017MaskR-CNN,Massa2018Maskrcnn-benchmark:PyTorch} is used to generate segment proposals for all objects in the scene. The network predicts segmentation mask along with the category label of each object. This is in contrast to NS-VQA, where colour, material, size and shape are predicted simultaneously, which is possible for CLEVR as all combinations of attributes generate only $96$ labels. In our SHOP-VRB, all combinations give $155520$ different labels, hence, only the category is predicted. Detected object masks in training and validation sets are verified with ground truth labels by choosing the intersection over union to be greater than 0.7. Next, the masked images of objects are composed with full scene images and input to ResNet-34 \cite{He2016DeepRecognition} to extract their properties along with 3D coordinates {(used for relative comparisons in cardinal directions)}, as shown in Figure \ref{fig:model}. This two-step approach results in a disentangled semantic representation of the visual input.

\paragraph{Description parsing} Our proposed text parser provides knowledge representation that is not observed in the image. All text descriptions are firstly tokenised and encoded using pretrained Glove \cite{Pennington2014GloVe:Representation} and then
parsed through bidirectional LSTM \cite{Hochreiter1997LongMemory} serving as an encoder. The hidden state of the encoder is concatenated with the one-hot vector representing detected visual attributes and parsed through a linear layer to obtain a list of text attributes. The training is performed in a fully supervised way with the use of multi-label classification loss (Binary Cross Entropy).

\paragraph{Question parsing} Attention seq2seq model \cite{Sutskever2014SequenceNetworks} with Bidirectional LSTM is used for both encoder and decoder parts of the question parser. The encoder outputs question embedding that is used to initialise the decoder and generate the symbolic program corresponding to the question. The training of question parser follows a two-step process. Firstly, the model is trained on a small group of question-program ground truth pairs with direct supervision. Then, the training is performed using REINFORCE \cite{Williams1992SimpleLearning} algorithm to tune the parser on the whole training set. Reinforcement part uses only the correctness of the answer generated by the symbolic program executor as the reward signal.
\paragraph{Program execution} It follows the pattern of  program executor from NS-VQA  \cite{Yi2018Neural-SymbolicUnderstanding} while expanding its domain for new types and values of object properties. Each module of the executor corresponds to a program-'value input' pair (if exists) of the ground truth programs (\eg  'filter\_color[yellow]' or only 'same\_size').
\section{Experiments}

We propose the following training regime to perform experiments on SHOP-VRB. For \textit{scene parsing} step, we train Mask R-CNN on training split for 30000 iterations. Subsequently, we train ResNet-34 for visual attributes extraction using training split and applying the stopping criterion based on validation split. \textit{Description parsing} LSTM is trained similarly. 

Supervised stage of \textit{question parsing} network is trained on a subset of questions selected from training split by sampling $1$ or $2$ questions from each question template for the visual part of the dataset only. 
The experiments with the visual part of data only are used for comparison to other methods and are followed by presenting the baseline results for vision and text data.
During testing, we use the test and the benchmark splits of the dataset, whereas most other works based on CLEVR use validation split for reporting results as the ground truth labels in the test split are not available. 

\subsection{Objects segments proposals}

We assess the performance of the segment proposal  using bounding box evaluation protocol from PASCAL VOC \cite{Everingham2010TheChallenge} and report the following mAP scores:
\begin{itemize}
    \item CLEVR (based on CLEVR mini from NS-VQA): \textbf{mAP=$1.0$},
    \item SHOP-VRB test split: \textbf{mAP=$0.9954$},
    \item SHOP-VRB benchmark split: \textbf{mAP=$0.4498$}.
\end{itemize}
It is clear that simple geometrical shapes of CLEVR pose no difficulty for segment proposal network. Moreover, based on the test split of SHOP-VRB we observe that recognising previously observed 3D models, even at a different position, orientation, material or colour is still an easy task, solved nearly perfectly. However, when challenged with new 3D shapes of objects belonging to the same category  the performance of the network drops drastically. 


\subsection{Attributes extraction}

We propose to analyse the performance of attribute recognition by breaking it down into separate properties (Table \ref{tab:attributes}).
\begin{table}[h]
\small
\begin{center}
\begin{tabular}{m{0.15\textwidth}>{\centering\arraybackslash}m{0.055\textwidth}>{\centering\arraybackslash}m{0.055\textwidth}>{\centering\arraybackslash}m{0.055\textwidth}>{\centering\arraybackslash}m{0.055\textwidth}}
\hline
Property    & test & B 0.7 & B 0.5  & B GT\\
\hline
Objects found       & $89.2$     & $69.9$     & $79.6$     &  $100$    \\
Correct category    & $88.3$     & $43.2$     & $47.4$     &  $58.5$    \\
Correct size        & $88.9$     & $61.2$     & $70.5$     &  $88.2$    \\
Correct weight      & $88.9$     & $53.7$     & $62.6$     &  $79.4$    \\
Correct colour      & $88.5$     & $50.8$     & $55.2$     &  $69.5$    \\
Correct material    & $88.4$     & $48.8$     & $54.3$     &  $71.6$    \\
Correct mobility    & $89.2$     & $65.4$     & $74.9$     &  $93.8$    \\
Correct shape       & $88.7$     & $38.0$     & $43.8$     &  $57.0$    \\
Overall             & $88.7$     & $51.4$     & $58.2$     &  $73.8$    \\
Mean distance error   & $0.062$     & $0.102$     & $0.125$     &  $0.153$    \\
\hline
\end{tabular}
\end{center}
\caption{Accuracy of property recognition for the visual part of SHOP-VRB (CLEVR not included as all scores were nearly perfect for all properties $\approx99.9$). Test and benchmark (B) columns correspond to splits of SHOP-VRB. Values next to B refer to thresholds on considering the detected mask as correct (based on IoU). Note that threshold 0.5 allows more masks to go through attribute recognition hence higher scores.  GT refers to the use of ground truth masks i.e., perfect segmentation, instead of Mask R-CNN predictions and provides the upper bound for the accuracy of the attributes recognition.} \label{tab:attributes}
\vspace{-2mm}
\end{table}
The experiments are performed on test and benchmark splits of our dataset. 
Firstly, we measure the percentage of objects segmented correctly following NS-VQA suggestion of recognising the match when the intersection over union of predicted and ground truth masks is larger than $0.7$ (we also experiment with $0.5$ for benchmark split). This assures that we consider only correctly recognised objects for attributes extraction limiting the scene representation assessment to items we have proper perception of.  Table \ref{tab:attributes} shows the percentage of found objects along with the accuracy of recognising each property,
 the overall accuracy of attribute extraction and the mean error for the 2D position of the object centre; the value is provided without units with $1.3$ corresponding to a mug height. All reported numbers are correctly recognised fractions of the total number of objects.

We observe that for the objects shapes that were observed during training, i.e. test split,  the recognition accuracy is approx. $89\%$. However, there is a clear performance drop from the test to the benchmark split, which contains different object shapes. Moreover, recognising the category and shape of new objects seems to be the most challenging task, whereas finding size and mobility is relatively easier. 
We observe that position estimation is worse for the benchmark split. However, there is an increase in the positioning error between experiments with different IoU thresholds. Allowing less accurate detections may improve property recognition  but slightly degrades the inferred position.  The use of ground truth masks also degrades the localisation performance, seemingly because of including the hardest samples into the experiment.


Table \ref{tab:attributes_textual} presents the recognition accuracy of textual properties.
\begin{table}[h]
\small
\begin{center}
\begin{tabular}{m{0.15\textwidth}>{\centering\arraybackslash}m{0.055\textwidth}>{\centering\arraybackslash}m{0.055\textwidth}>{\centering\arraybackslash}m{0.055\textwidth}>{\centering\arraybackslash}m{0.055\textwidth}}
\hline
Property    & T-GT & T-A & B-GT & B-A\\
\hline
Correct disassembly    & $78.6$     & $72.0$     & $77.3$     &  $49.7$    \\
Correct picking up        & $74.6$     & $67.2$     & $57.7$     &  $41.6$    \\
Correct powering      & $90.0$     & $79.3$     & $87.4$     &  $58.4$    \\
Multi-label avg F1      & $0.41$     & $0.41$     & $0.33$     &  $0.25$    \\
\hline
\end{tabular}
\end{center}
\caption{Accuracy of textual property recognition for SHOP-VRB (test (T) and benchmark (B) splits). GT refer to the use of ground truth attributes as inputs to description parsing network, whereas A correspond to the use of outputs form attributes extraction network.} \label{tab:attributes_textual}
\vspace{-2mm}
\end{table}
\begin{table*}[!t]
\small
\begin{center}
\begin{tabular}{m{0.23\textwidth}>{\centering\arraybackslash}m{0.095\textwidth}>{\centering\arraybackslash}m{0.095\textwidth}>{\centering\arraybackslash}m{0.095\textwidth}>{\centering\arraybackslash}m{0.1\textwidth}>{\centering\arraybackslash}m{0.1\textwidth}>{\centering\arraybackslash}m{0.095\textwidth}}
\hline
Split    & Count & Exist & Compare Number  & Compare Attribute & Query Attribute & Overall\\
\hline
CLEVR (NS-VQA)      & $99.7$     & $99.9$     & $99.9$     & $99.8$     & $99.8$     & $99.8$     \\
Test - q1/q2           & $84.4$/$94.4$     & $92.0$/$96.8$     & $58.8$/$96.2$     & $94.0$/$93.6$     & $89.5$/$93.5$     & $88.0$/$94.2$ \\
Benchmark - q1/q2      & $35.5$/$36.1$     & $59.8$/$59.2$     & $50.2$/$62.3$     & $53.3$/$53.2$     & $27.8$/$26.6$     & $38.7$/$38.8$ \\
Benchmark GT - q1/q2   & $44.4$/$46.5$     & $64.5$/$65.9$     & $51.4$/$68.4$     & $56.6$/$56.0$     & $35.7$/$35.7$     & $45.2$/$46.6$ \\
\hline
Test GT attributes   & $57.9$/$82.1$     & $73.7$/$90.5$     & $76.7$/$85.2$     & $40.2$/$81.8$     & $45.1$/$81.5$     & $55.8$/$83.0$ \\
Test recovered attributes  & $47.9$/$67.5$     & $69.3$/$83.1$     & $74.0$/$80.0$     & $29.0$/$70.1$     & $39.5$/$67.2$     & $55.8$/$49.1$ \\
Benchmark GT attributes   & $52.7$/$80.1$     & $70.4$/$89.2$     & $71.2$/$82.0$     & $39.0$/$81.7$     & $42.0$/$80.4$     & $51.7$/$81.7$ \\
Benchmark recovered attributes   & $23.8$/$29.7$     & $53.0$/$57.0$     & $60.0$/$59.4$     & $3.0$/$41.0$     & $23.2$/$25.0$     & $29.8$/$35.0$ \\
\hline
XNM GT/GT \cite{Shi2019ExplainableGraphs}      & $98.7/98.0$     & $99.5/99.5$     & $99.7/98.8$     & $99.9/98.7$     & $99.5/99.5$     & $99.6/99.0$     \\
FiLM \cite{Perez2018FiLM:Layer}          & $65.7/51.6$     & $82.1/68.5$     & $75.7/72.0$     & $85.3/68.9$     & $91.2/63.1$     & $83.5/62.1$ \\
MAC \cite{Hudson2018CompositionalReasoning}          & $66.0/51.0$     & $81.9/67.9$     & $79.7/71.8$     & $80.0/68.0$     & $88.3/61.9$     & $81.4/61.2$ \\
TbD \cite{Mascharka2018TransparencyReasoning}      & $48.1/42.8$     & $61.4/55.3$     & $70.5/70.3$     & $73.8/66.2$     & $62.0/49.0$     & $62.0/52.9$ \\
RN \cite{Santoro2017AReasoning}     & $52.3/22.8$     & $70.5/29.4$     & $69.5/35.8$     & $70.3/34.2$     & $75.4/22.1$     & $69.3/26.0$ \\
\hline
\end{tabular}
\end{center}
\caption{Question answering accuracy.  (Top) NS-VQA results on CLEVR benchmark provide the baseline. We experiment on SHOP-VRB test and benchmark  using $1$ or $2$ samples of each template for the supervised training and also provide the results for attributes inference based on perfect segmentation (GT) for the visual part of the dataset. 
 (Middle) The results for  textual questions accuracy/all data accuracy with the same settings as the text parsing module and q2 sampling. These results show that our dataset provides a challenging benchmark for visual  reasoning.
(Bottom) The results for state of the art VQA methods on SHOP-VRB. Scores are reported as: \textit{test}/\textit{benchmark} accuracy for the given question types in columns. GT/GT symbol for XNM method indicates using both ground truth scenes and ground truth programs. 
} \label{tab:questions}
\vspace{-3mm}
\end{table*}
We show the scores for test and benchmark splits for two scenarios: providing the description parsing with ground truth visual attributes or using the ones extracted by scene parsing. When using the ground truth labels the recognition challenge posed by the benchmark split is removed, which leads to comparable test and benchmark scores. However, when applying the full pipeline, errors in visual recognition drastically affect the recognition of textual properties. We also observe that increase in the number of possible labels lowers the recognition score (2-class \textit{powering} is the easiest to classify, whereas multi-label choice poses a greater challenge).

\subsection{Question answering}

The accuracy of the question answering task is presented in Table \ref{tab:questions}. 
We follow the division for question types proposed in CLEVR. We experiment using CLEVR  and NS-VQA \cite{Yi2018Neural-SymbolicUnderstanding}, as well as the test and the benchmark splits of our dataset. We use both inferred and ground truth masks for obtaining scene representations, and testing both sampling schemes of $1$ and $2$ questions per template for the supervised part of training.

We observe similar behaviour of question answering to the results for attributes recognition. As a consequence of the increased difficulty in obtaining accurate scene representation for the benchmark split of SHOP-VRB, the answers accuracy also drops drastically. There is no considerable decrease in performance when using $2$ times fewer questions for supervised training and benchmark testing.
On the other hand, the test split shows a considerable difference in performance in that case, especially for questions concerning comparing  integer quantities, which suggests insufficient reward signal from those questions during the reinforcement phase. 
In benchmark results,  the lowest performance is observed when asking about particular properties of objects, which is due to having inaccurate semantic scene description. The best accuracy can be observed for the easiest to guess questions \eg querying about the existence of a certain object or asking to compare two numbers inferred from the scene. 

Overall, we observe a large drop in the attributes recognition, question answering and reasoning  when we evaluate them on new object shape instances. More difficult task exposes also inaccuracies of multi-stage approaches due to accumulation of errors.

When the new task of recovering additional properties from the text is introduced, we observe a significant performance decrease. Another step of reasoning (description parsing) relying on the intermediate output of the model (visual attributes) introduces additional error.  The text parsing is equally challenging for both test and benchmark splits causing the decrease in overall question answering accuracy. We also observe text-based questions to be the major cause of the performance drop. We argue that multi-label classification poses a significant challenge for the model, hence, the program executor is easily  misled.

In summary, embedding textual descriptions into the model introduces new and significant challenges, but enables a new source of knowledge for extracting information not visible in images.

\subsection{VQA and reasoning experiments}

The experiments for the question answering accuracy were performed for several state of the art VQA  methods. All these methods were evaluated with CLEVR  and obtained nearly perfect scores  (over $95\%$ in all cases). We use the original implementations and parameters to carry out experiments with the proposed SHOP-VRB dataset. The results for both test and benchmark splits of SHOP-VRB are shown in Table \ref{tab:questions} (Bottom).
We observe that XNM  \cite{Shi2019ExplainableGraphs}, that  executes functions on graph nodes, performs nearly perfectly on both splits when ground truth scenes and ground truth programs are used. This confirms that the question-answer pairs were prepared correctly and there are no non-answerable queries. 
In addition, with supervision, the programs are recovered from questions with over $99.9\%$ accuracy using simple Seq2Seq \cite{Johnson2017InferringReasoning}. For RN \cite{Santoro2017AReasoning}  trained on images only we observe little generalisation as the results for \textit{test} and \textit{benchmark} are significantly different. RNs present the simplest network model thus limiting its more complex reasoning capability. 
FiLM \cite{Perez2018FiLM:Layer}, MAC \cite{Hudson2018CompositionalReasoning} and TbD \cite{Mascharka2018TransparencyReasoning} present much better results for \textit{benchmark} split, indicating better generalisation than NS-VQA. We can infer that keeping the latent representation not fully disentangled enables the network to use visual cues that are typically not used by the human for abstract reasoning. 
However, it is seemingly harder to reason about known objects when the representation is entangled  by comparing to NS-VQA  \textit{test} accuracy. The aforementioned approaches present the reasoning chain as a flow of attention over the image. In contrast to NS-VQA, these methods lack full transparency in object representation. Hence, we argue that SHOP-VRB is suitable for evaluating the quality of disentangled scene representation while being very challenging for the VQA approaches. 
\section{Conclusions}

In this work, we proposed an approach and a  more realistic benchmark for experiments in robotics object perception and visual reasoning than the existing datasets, while introducing a new task of multimodal reasoning. 
We demonstrated that testing a method on a set of previously observed 3D shapes is insufficient to assess the generalisation properties of reasoning systems. Our experiments demonstrate that the task becomes  challenging if more complex object classes and different instances of objects are used in the evaluation. SHOP-VRB also exposed the weakness of a multi-stage reasoning approach by showing the accumulation of errors from critical stages of this method. 
The introduction of additional data source such as natural language with a corresponding approach for extracting text-based object attributes is also shown to be an interesting challenge while greatly expanding the task of reasoning. This allows for use of multimodal data making it more  grounded in the robotics applications.  
We believe that the proposed approach and the dataset can  bridge the gap between robotics visual reasoning and question answering.



\hfill


\noindent{\bf Acknowledgements.}
 This research was supported by UK EPSRC IPALM project EP/S032398/1.
\bibliography{references.bib}

\end{document}